# Interactively Diagnosing Errors in a Semantic Parser

**Constantine Nakos**  CNAKOS@U.NORTHWESTERN.EDU
**Kenneth D. Forbus**  FORBUS@NORTHWESTERN.EDU
Qualitative Reasoning Group, Northwestern University, 2233 Tech Drive, Evanston, IL 60208 USA

## Abstract

Hand-curated natural language systems provide an inspectable, correctable alternative to language systems based on machine learning, but maintaining them requires considerable effort and expertise. Interactive Natural Language Debugging (INLD) aims to lessen this burden by casting debugging as a reasoning problem, asking the user a series of questions to diagnose and correct errors in the system's knowledge. In this paper, we present work in progress on an interactive error diagnosis system for the CNLU semantic parser. We show how the first two stages of the INLD pipeline (symptom identification and error localization) can be cast as a model-based diagnosis problem, demonstrate our system's ability to diagnose semantic errors on synthetic examples, and discuss design challenges and frontiers for future work.

## 1. Introduction

Hand-curated natural language systems[1] provide an inspectable, correctable, and incremental alternative to systems based on machine learning (ML). Where ML models are opaque and provide limited opportunities for debugging or incremental extension, hand-curated language systems can be improved by editing their linguistic knowledge. Such changes are typically local, do not require retraining, and can be verified by both testing and direct inspection.

The downside of hand-curated language systems is that they are difficult to maintain. Their knowledge spans the breadth of language itself, at least in principle, while it is stored in a variety of specialized formats, such as grammar rules or lexicon entries. Traces of a language system's behavior are complicated and can stump even experts. More generally, the burden of decision-making falls on the maintainer, rather than the system, making maintenance costly. Even when parts of the system's knowledge are learned automatically, the maintainer is responsible for the design, representation decisions, and knowledge curation that keep the system running.

*Interactive Natural Language Debugging* (INLD; Nakos, Kuthalam, & Forbus, 2022) aims to lessen this burden. INLD is a framework for debugging hand-curated language systems by asking the user questions in natural language and reasoning about the answers. With the right questions, the system can determine where its behavior deviates from the user's expectations, locate the error in its knowledge that caused the discrepancy, and repair it with the user's assistance.

In this paper, we present an implementation of the first two stages of the INLD pipeline: symptom identification and error localization. Our interactive error diagnosis system uses

---
[1] Those with explicit linguistic knowledge maintained by a human.

```
What part of speech is "bob"?
1) noun
2) proper noun

(Please enter a list of numbers between 1 and 2, or "none".)
> 2

What does "wedge" mean?
1) wedge (a golf club)
2) wedge shaped thing (a geometrical figure)

(Please enter a list of numbers between 1 and 2, or "none".)
> none

These assumptions are faulted:
 -Choice Set #8748 ("wedge") is complete.
```

*Figure 1*. Sample diagnosis transcript for the sentence "Bob ate the wedge."

techniques from model-based diagnosis to debug semantic errors in CNLU (Companions Natural Language Understanding; Tomai & Forbus, 2009), a parser that relies on a broad range of hand-curated knowledge to produce rich semantic interpretations of a piece of text. A sample transcript is shown in Figure 1. While work on the system is ongoing, its current state is sufficient to illustrate how INLD can be framed as a model-based diagnosis problem, how decomposition strategies allow the system to drill down on specific errors, and how natural language questions let a user diagnose errors without specific expertise.

The rest of the paper is as follows. We begin by reviewing relevant background on CNLU and the INLD pipeline. Next, we describe our approach to diagnosis, including how we formulate the initial reasoning problem and decompose it in response to the user's answers. This is followed by a demonstration of our system's ability to diagnose semantic errors in synthetic examples. We conclude with a discussion of related work, closing thoughts, and frontiers for future work.

## 2. Companions Natural Language Understanding

CNLU (Tomai & Forbus, 2009) is the language understanding component for the Companion cognitive architecture (Forbus & Hinrichs, 2017). It parses text to produce a set of parse trees and semantic interpretations, the latter of which are grounded in the NextKB ontology.[2] The parsing algorithm is based on Allen's (1994) bottom-up chart parser and has been extended to build up semantic as well as syntactic structure.

The primary function of CNLU is to supply semantic interpretations for downstream planning and reasoning tasks. It has been instrumental in allowing Companions to learn norms (Olson & Forbus, 2021), games (Hinrichs & Forbus, 2014), commonsense knowledge (Ribeiro & Forbus, 2021), and qualitative knowledge (Crouse, 2021), and it serves as the backbone for a question-answering information kiosk located in Northwestern's CS department (Wilson et al., 2019).

There are four types of knowledge used by CNLU, corresponding to the lexicon, grammar, semantics, and ontology:

---

[2] http://www.qrg.northwestern.edu/nextkb/index.html

- *Lexicon entries* map the surface form of a word to its root, part of speech, and grammatical features. For example, the token `ate` maps to the root `Eat-TheWord` as a past-tense verb.
- *Grammar rules* govern how smaller syntactic constituents can be composed into larger ones. They use grammatical features to enforce rules such as subject-verb agreement.
- *Semtranses*[3], derived from FrameNet (Fillmore et al., 2001), map words to predicate calculus templates that encode their meanings in NextKB. *Valence patterns* track the valid combinations of syntactic and semantic roles for a semtrans.
- *Type information*, a subset of the NextKB ontology, helps CNLU filter out semantically incoherent interpretations.

While a large part of this knowledge was derived from existing resources, it is now maintained manually by the developers of CNLU. The specialized knowledge formats and intricacies of the parser limit the pool of maintainers, making CNLU a prime testing ground for INLD.

One important detail is the way CNLU handles ambiguity. Because CNLU is designed for use within a larger reasoning pipeline, it adopts a wait-and-see approach to disambiguation. If a sentence is syntactically or semantically ambiguous, CNLU will produce a set of *choice sets*, each of which consists of a set of mutually exclusive and exhaustive *choices* which represent different meanings of a word or different parse trees for the sentence. Choice sets are a compact way to encode a large space of potential interpretations. They are the grist for the domain-specific reasoning processes used to produce a final interpretation.[4]

Figure 2 shows the semantic choice sets for a sample sentence "Bob ate the wedge." in the CNLU interface. The upper row shows the choices for "ate", and the lower shows the ones for "wedge". `eat4200`, `bob4152`, and `wedge4237` are *discourse variables* representing the entities the text refers to. *Role relations* like `performedBy` and `consumedBy` connect events to their arguments, a form of Neo-Davidsonian representation (Parsons, 1990). The grey boxes show the FrameNet frame of the semtrans the choice was derived from. Note that CNLU only knows two senses of the word "wedge" that grammatically fit the sentence: the kind of shape (`Wedge`) and the kind of golf club (`Wedge-GolfClub`). The intended meaning, a kind of sandwich, is missing.

As a final point, choices in CNLU can have dependencies. For example, it makes little sense to select the golf club sense of "wedge" with a parse tree that treats it as a verb, if one exists. Likewise, the choices for adjectives depend on the noun they modify (e..g., `(LargeFn Cat)` for "large cat", where `Cat` is derived from the noun's choice set). The `enablesChoice` relation captures this dependency. A choice can only be selected if at least one of the choices that enables it is still in play, enforcing consistency across selected choices.

## 2.1 Error Taxonomy

Table 1 shows a taxonomy of error types found in CNLU's knowledge, along with the symptoms that can be used to diagnose them. Some symptoms can be identified automatically by the parser (e.g., a fragmented parse), while others require user judgments to discern (e.g., a missing word sense). The errors correspond to CNLU's four types of knowledge, and they have varying effects on the parse depending on how the knowledge is used.

---

[3] "Semantic translations."
[4] Such processes are heterogenous and may involve considerable world knowledge, placing them beyond the scope of INLD. Instead, we try to ensure that at least one valid interpretation for a sentence exists.

Figure 2. Semantic choice sets for the sentence "Bob ate the wedge."

It is worth distinguishing between *errors of restriction* and *errors of permission*.[5] Errors of restriction occur when the language system's knowledge is too restrictive to produce a desired interpretation. For example, a missing valence pattern might prevent CNLU from linking the right sense of a verb to its arguments, in which case that verb sense would be dropped. Errors of permission occur when the language system's knowledge is too permissive and allows nonsensical interpretations. For example, a semtrans that maps the word "lack" to the concept `OwningSomething` will be incorrect in all sentences and should be excised if possible. Errors of restriction are marked in **bold** in Table 1, and errors of permission are marked in *italics*.

Errors of restriction are much easier to diagnose than errors of permission because they only involve the sentence at hand. In contrast, errors of permission require looking at the implications of a piece of knowledge in all possible contexts. For example, the conventional interpretation of "Mary had a little lamb." does not allow the `BirthEvent` sense of the word "had", but that sense is preferred in the sentence "Mary had a child." Even though `BirthEvent` seems erroneous in the first sentence, flagging it as an error in CNLU's knowledge would be wrong.

Diagnosing errors of permission requires more sophisticated debugging techniques, such as generating sentences to reveal the system's misconceptions (e.g., "The man lacked the cat." when he should own it). Fortunately, errors of permission do not affect a parser's ability to produce the right interpretation. They merely add incorrect interpretations for the system to filter through. Thus, for the purposes of this paper, we focus our attention on errors of restriction.

## 3. Interactive Natural Language Debugging

Interactive Natural Language Debugging (INLD) is a framework for debugging hand-curated language systems by asking the user a series of questions in natural language. The answers allow the system to diagnose and correct errors in its linguistic knowledge, removing the need for an expert to crawl through a parse and make the correction manually.

---

[5] By analogy with "errors of omission" and "errors of commission". In principle, INLD encompasses both understanding and generation, so we define our own terms to avoid confusion.

*Table 1*. Taxonomy of types of errors found in CNLU

| ID | Error | Category | Symptoms |
|----|-------|----------|----------|
| **A1** | **Missing lexicon entry** | **Lexicon** | **Identified by parser** |
| *A2* | *Incorrect lexicon entry* | *Lexicon* | *Fragmented parse or incorrect parse tree* |
| **B1** | **Missing grammar rule** | **Grammar** | **Fragmented parse or incorrect parse tree** |
| *B2* | *Incorrect grammar rule* | *Grammar* | *Fragmented parse or incorrect parse tree* |
| **C1** | **Restrictive type constraint** | **Semantics** | **Type checking rules out correct interpretation** |
| **C2** | **Missing valence pattern** | **Semantics** | **Valence pattern checking rules out correct interp.** |
| **C3** | **Missing word sense** | **Semantics** | **Identified by parser or no correct interp. at all** |
| *C4* | *Permissive type constraint* | *Semantics* | *Bad interp. allowed or generates bad paraphrase* |
| *C5* | *Incorrect valence pattern* | *Semantics* | *Bad interp. allowed or generates bad paraphrase* |
| *C6* | *Incorrect semantics* | *Semantics* | *Correct interp. missing or generates bad paraphrase* |
| **D1** | **Missing inheritance info** | **Ontology** | **Type checking rules out correct interpretation** |
| *D2* | *Incorrect inheritance info* | *Ontology* | *Bad interp. allowed or generates bad implication* |
| *D3* | *Missing disjointness info* | *Ontology* | *Bad interp. allowed or generates bad implication* |
| **D4** | **Incorrect disjointness info** | **Ontology** | **Type checking rules out correct interpretation** |

The intuition is straightforward. Humans are fluent in language and have rich commonsense knowledge that the system lacks, but they may not have the time or understand the system well enough to debug it. With the right setup, the language system can reason about its own internals, but it does not have the commonsense knowledge to know whether it has interpreted a sentence correctly. By posing diagnostic questions to the user and reasoning about the answers, the system can home in on errors and perform its own debugging. Thus, INLD takes advantage of the user's knowledge for debugging while insulating the user from the complexities of the system.

INLD is a pipeline with four stages, as shown in Figure 3. The input is a sentence suspected to contain an error,[6] and the output is a correction to the system's linguistic knowledge.

The first stage of INLD is *identification*, which takes an input sentence, parses it, and identifies a set of *symptoms* that characterize the potential error, such as the ones shown in Table 1. Symptoms guide debugging and serve as its success criteria. If a proposed correction fixes the symptoms without introducing new ones, debugging has succeeded. Otherwise, the system may need to backtrack or initiate another round of debugging.

The second stage of the pipeline is *localization*, which takes a set of symptoms and attempts to locate the error that caused them. Localization uses several strategies depending on the nature of the symptom, but the general pattern is to drill down through a combination of introspection (e.g., looking for word senses that were ruled out) and questions to the user (e.g., whether a ruled-out sense would have been acceptable). The output is a *diagnosis*, a piece of incorrect or missing knowledge that explains the observed symptoms.

---

[6] The question of efficiently finding such sentences is an interesting research question in its own right. For now, we treat INLD as a tool to be invoked when an error is suspected, not one that will seek errors out.

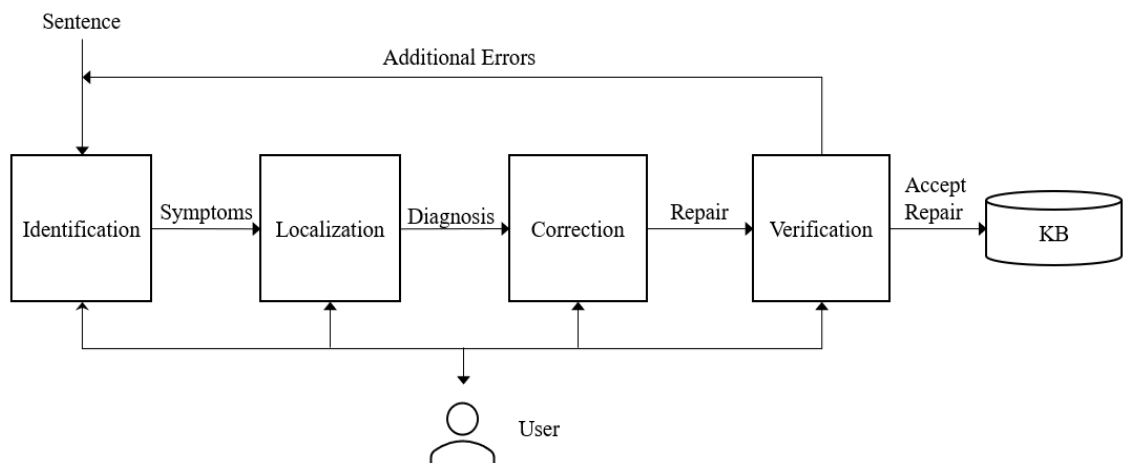

*Figure 3*. The INLD Pipeline.

After localization comes *correction*, where the system works with the user to repair the error diagnosed in the previous stage. While some corrections are straightforward, others require significant expertise, placing them outside the reach of INLD. In principle, lexicon entries, semtranses, or valence patterns can all be repaired by a novice user if the system asks the right questions. But grammar rules are more complex and may require expert intervention.

The final stage of the pipeline is *verification*, which takes a proposed repair and either accepts it, integrating it into the system's knowledge, or rejects it, kicking off another round of INLD. Verification can be *local*, confirming that the repair works for the current sentence, or *global*, testing the change against other sentences to look for any side effects.

Notably, INLD is not an all-or-nothing proposition. Even partial information about an error may be enough to kick-start manual debugging by an expert, making the process worthwhile even if it is incomplete. Enabling novice users to debug common errors also frees up experts to focus on more challenging problems. So, while we aspire to reach complete coverage with our INLD system, perfection is not necessary to make meaningful progress.

For the remainder of this paper, we focus our attention on the first two stages of the pipeline, symptom identification and error localization, in the context of CNLU. As we will see, these two stages fit naturally into a model-based diagnosis framework.

## 4. Interactive Error Diagnosis

While there are a number of ways to implement the identification and localization stages of the INLD pipeline, we select a *model-based diagnosis* framework. Model-based diagnosis compares the behavior of an *artifact* with the predictions of a *model*, using the discrepancies to determine where the artifact is malfunctioning (taking the model as ground truth) or where the model fails to capture its behavior (taking the artifact as ground truth). The artifact can be physical, like a printer, or conceptual, like a student's reasoning on a homework problem.

For the purposes of INLD, the artifact is the user's understanding of a sentence, and the model is a trace of the parser's behavior as it tries to interpret the same sentence. The user's understanding is taken as ground truth. Any incompatibilities with the parser's interpretation indicate a *fault* in the model, corresponding to a gap or error in the parser's knowledge.

Naturally, the user's understanding of a sentence cannot be examined directly, so the system must ask the user questions to capture relevant aspects of it, such as the user's judgments about word senses or parts of speech. These questions are analogous to the *measurements* used to diagnose faults in physical systems. The same way that checking the output of various parts of a printer can be used to triangulate which one is broken, targeted questions can be used to determine where the user's understanding and the parser's understanding differ.

In the rest of this section, we describe an implemented diagnosis engine for CNLU, which currently supports semantic error diagnosis. We begin by reviewing GDE, continue by discussing our problem formulation and diagnosis algorithm, and conclude with decomposition strategies.

**4.1 GDE & CATMS**

The backbone of our diagnosis system is the General Diagnostic Engine (GDE; de Kleer & Williams, 1987). GDE is a model-based diagnosis algorithm that eschews domain-specific diagnosis strategies in favor of a general approach. Given a model and the set of *observations* made so far, GDE computes the measurement to take that will best distinguish between the current *minimal diagnoses*, minimal sets of faults that explain the observed discrepancies between the artifact and the model. The algorithm terminates when only one minimal diagnosis remains or when no more measurements are possible.

GDE uses an assumption-based truth maintenance system (ATMS; de Kleer, 1986) to keep track of its hypotheses. An ATMS maintains a set of *nodes*, corresponding to assertions, whose *labels* compactly encode the logical *environments* in which they are true. Each environment consists of a set of *assumptions*, nodes assumed to be true in any environment they appear in. *Justifications* connect their antecedents to a consequent, combining the antecedents' labels to determine when the consequent is true. Some nodes are marked as *contradictions*. Environments that entail a contradiction are marked *nogood* and removed from further consideration. These building blocks allow an ATMS to reason about all combinations of assumptions at once, letting it track large sets of hypotheses at the risk of exponential blow-up in the size of its labels.

To keep the size of the ATMS tractable, we use a CATMS (DeCoste & Collins, 1991), which *compresses* its labels by blocking propagation when an assumption is reached. The label of an assumption will only ever contain the singleton environment corresponding to itself, so just one environment is passed to downstream nodes, rather than the exponential number of environments needed to list all the ways the assumption could be entailed. The tradeoff is that the CATMS has to do more work at runtime, recomputing some of the environment comparisons that an ordinary ATMS caches in its (uncompressed) labels. Even so, the runtime benefits of smaller labels typically outweigh the extra work needed to unpack them.

In practical terms, this means we can prevent the combinatorial blow-up of labels by turning select nodes into assumptions. Effectively, we are carving the problem into separable pieces that use assumptions to isolate the complexity of one part of the model from the rest of it. As we will show, our INLD formulation lends itself to this technique, with natural groupings of nodes whose internal behavior can be summarized with a single assumption. We have found that the efficiency gains provided by a CATMS are crucial to having our system run in real time.

### 4.2 Problem Formulation

*4.2.1 The Algorithm*

With our implementation of GDE in place and a CATMS to keep label sizes tractable, we turn our attention to the diagnosis problem itself. We divide our algorithm into an *inner loop* and an *outer loop*, as shown in Figure 4. The inner loop is our implementation of the GDE algorithm, asking the user a series of questions to diagnose faults in the model. Once the inner loop has found a likely fault, the outer loop *decomposes* the model, reformulating the problem to drill down on the fault and refine the diagnosis.

For example, a missing sense for a word might cause the algorithm to look for senses of that word that were (perhaps erroneously) ruled out during parsing and add them to the model for consideration. The inner loop then takes over to ask further questions, in this case to determine if any of the recovered senses is the right one and, if so, why it was ruled out. Depending on the fault that the algorithm finds, another round of decomposition may be necessary to locate the error (e.g., an overly restrictive type constraint). The cycle continues until the algorithm finds a diagnosis that cannot be decomposed, corresponding to underlying errors in CNLU's knowledge.

There are three reasons to structure our algorithm this way. The first is *tractability*. While the CATMS helps keep the runtime of the algorithm manageable, the size of the problem is still a significant concern. Having to instantiate the parse trace in its entirety is wasted effort when the problem can be decomposed into smaller pieces.

The second reason is *control*. The outer loop allows us to control when structure is added to the model and how it is explored. We can carve off manageable pieces of massive search spaces (e.g., the set of rules that did *not* fire during the parse), backtrack when a decomposition strategy does not pan out, and instantiate structure dynamically to support flexible debugging strategies (e.g., differential diagnosis; Nakos et al., 2022).

The final reason for the structure of our algorithm is *modularity*. Decomposition strategies are an organic way to handle different types of errors and different techniques for localizing them. New strategies can be added to the library as they are formulated, allowing us to incrementally extend the capabilities of our algorithm. As stated in Section 3. , INLD is not an all-or-nothing proposition, and nowhere is that more evident than in decomposition.

Next, we turn to the model itself and the key concepts that will help us transform a parse trace into a model our diagnosis algorithm can use.

*4.2.2 Completeness Assumptions*

The challenge before us is to formulate our model in such a way that fault diagnosis in the model leads to error diagnosis for CNLU. We must capture enough of the logical dependencies between different elements of the parse that the measurements our algorithm takes—the user's answers to its questions—will eventually localize the error. Faults in the model should either correspond to CNLU errors directly or represent elements of the parse that may contain errors.

The crux of the problem is deciding what to use as *defaults*: assumptions that are presumed to be true unless there is evidence that they are not. GDE computes the set of diagnoses by querying the ATMS for all consistent, maximal combinations of defaults. The defaults that are excluded from a combination are the ones that are inconsistent with it, indicating faults.

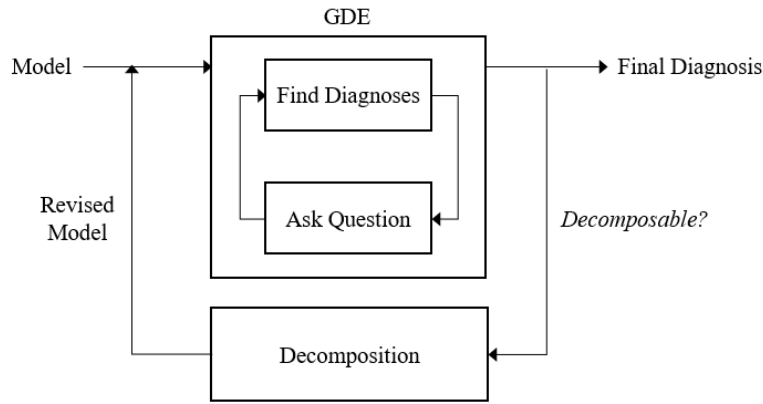

*Figure 4.* Diagnosis algorithm.

For an electrical circuit, the defaults might be assumptions that the circuit's components are operating properly. If the observed behavior of the circuit does not match the predicted behavior of the model, some of the defaults must be inconsistent with each other; at least one component must be faulted. Thus, the defaults excluded from a diagnosis are its faults. The GDE algorithm then filters the set of diagnoses to the ones with the fewest faults and proposes a measurement to take that can distinguish between the diagnoses that remain.

For CNLU, we use *completeness assumptions* as defaults. A completeness assumption is a type of closed-world assumption that states some set of parse elements is *complete*, meaning that one of the elements belongs in a valid interpretation of the sentence. For example, the choice set for the word "wedge" would be complete if one of the choices captured the semantics of the word in the context of the sentence, and it would be *incomplete* if the right sense was missing. There might also be completeness assumptions for the parse trees for the sentence, the semtranses for a word, or the valence patterns for a semtrans: anywhere an element might be missing.

While completeness assumptions are useful for linking model faults to CNLU errors, they do a poor job of driving diagnosis. They are weak assumptions—in essence, "one of the elements in this set is right"—that do not cause the conflicts the GDE algorithm needs to find faults.

Therefore, we pair each completeness assumption with an *incompleteness assumption*, which states that the set of parse elements is incomplete. This is a much stronger claim—"none of the elements in this set are right"—which gives the GDE algorithm something to work with. Making sure the ATMS knows that the completeness and incompleteness assumptions for the same set are contradictory, we add the incompleteness assumptions to the list of defaults, where they will be assumed alongside the completeness assumptions by the GDE algorithm. However, it is not meaningful for an incompleteness assumption to be faulted (i.e., it is good when a set of elements is complete), so we do not consider them to be faults when they are excluded from a combination.

### 4.2.3 Acceptability Judgments

To put our completeness assumptions to work, we must next define what it means for a parse element to be "right". This is complicated by two issues. First, sentences may be ambiguous, in which case there is no single "right" interpretation to find. For example, "Bob threw the wedge" is ambiguous as to the type of wedge, so it has more than one valid interpretation. Second, a piece of knowledge may be correct even if it does not apply to the current sentence. For example, we would not delete the golf club sense of the word "wedge" just because it is the wrong sense for "Bob ate the wedge."

To accommodate ambiguity and context dependence, we say a parse element is *acceptable* if it belongs to a desired interpretation of the sentence and *unacceptable* if it does not.[7] This helps the system reason about the parse. The acceptability of one element might depend on the acceptability of another, and this dependence can be captured with ATMS justifications. For example, the first semantic choice for the word "eat" in Figure 2 is only acceptable if all of its conjuncts are acceptable. Note that the other choice for "eat" shares two of the same conjuncts (`performedBy` and `consumedObject`), so the system can gather information about both choices by asking the user whether these expressions are acceptable.

Acceptability also gives us a way to formalize measurements. When the system asks the user a question—whether yes-or-no (e.g., "Is 'wedge' a noun in this sentence?") or multiple choice (e.g., "What does 'wedge' mean?")—the response it receives will be one or more *acceptability judgments* reflecting the user's verdict for those parse elements. With enough such judgments, the algorithm can determine which completeness assumptions are faulted, localizing the error.

Measurements are recorded in the algorithm's problem state and are always treated as true for the purpose of calculating diagnoses. For example, if all of the parse elements of a completeness assumption have been judged unacceptable, the assumption must be faulted; the set is incomplete.

Not all parse elements can be measured directly. Some are too complicated or too specialized to express to the user. For example, no one but the most experienced users will be able to tell if a parse tree is acceptable just by looking at it, but users with a basic understanding of grammar can tell if "wedge" is a noun, which might be sufficient to rule out several parse trees. Clever use of indirect measurements can open even some of the most baroque parse elements to user scrutiny.

### 4.2.4 Factored Interpretations

The last piece of the puzzle is to arrange the parse elements in a way that reflects the logic of the parse. Our goal is to have a *root element* that is acceptable if at least one acceptable interpretation of the sentence exists and unacceptable otherwise.

One naïve way to accomplish this would be to have a separate element for each interpretation (i.e., each consistent combination of CNLU choices) and connect them all to the root. This would get the job done—the label of the root would capture the full set of scenarios where an acceptable interpretation existed—but only at the cost of a combinatorially prohibitive number of nodes.

Another naïve approach would be to have a parse element for each choice set and connect them all jointly to the root. Each choice set would be acceptable when at least one of its choices was acceptable, and the root would be acceptable when all of the choice sets were. While this is a much more compact setup, it ignores the dependencies between choices. For example, if the only acceptable parse tree has "wedge" as a verb and the only acceptable sense of "wedge" requires it to be a noun, both choice sets would be satisfied even though no consistent interpretation exists.

Instead of either of these approaches, we use *factored interpretations* to correctly capture the space of possible interpretations by taking advantage of the dependencies between choices. The `enablesChoice` relation, which determines when choices should be considered, is purely local. Choices are enabled as long as none of the choices that directly enable them have been ruled out, and no other choices affect this. Therefore, the enablement graph encodes the space of possible interpretations. Every consistent interpretation can be found by selecting an enabled choice, ruling out any other choices in the same choice set (e.g., "wedge" cannot mean "shape" and "golf club" at the same time), and repeating until no choices are left.

---

[7] "Desired" gives us latitude to debug specific interpretations. Even if an interpretation is valid, it may not be the one we wish to focus on. Put another way, acceptability depends on the intentions of the user.

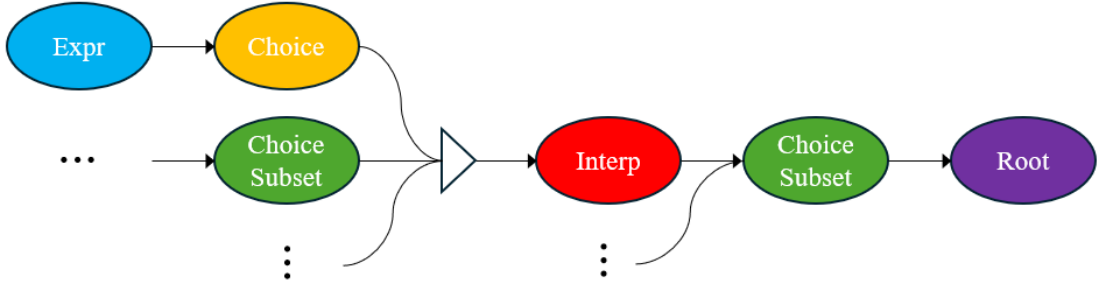

*Figure 5*. Simplified representation of a model generated by our algorithm. Each oval represents a parse element, stored as a pair of nodes in the ATMS (*acceptable* and *unacceptable*). Each arrow represents a justification. User acceptability judgments about, e.g., semantic expressions are installed as assumptions (blue), which in turn determine choice acceptability (orange). Factored interpretations (red) and choice subsets (green) encode the valid interpretations of the sentence, which the root (purple) aggregates.

Under this view, every choice represents a set of partial interpretations: all the ones that can be constructed by walking forward from that node in the graph. Because enablement is local, the partial interpretations constructed this way are independent of any prior choice. All that matters is *that* the initial choice is enabled, not *how* it is enabled.

This insight is the basis for our factored interpretations. The factored interpretation rooted at choice $C$ is defined recursively as a parse element that is acceptable if (a) $C$ is acceptable and (b) for each *choice subset* enabled by $C$, the factored interpretation rooted at one of those choices is acceptable. Choice subsets are groups of choices enabled by $C$ that belong to the same choice set.

Factored interpretations let us compactly encode when a parse is acceptable. Beginning with the fully independent choices (i.e., parse trees), we recursively install factored interpretations and choice subsets until all factored interpretations have been accounted for. The connections to each individual choice allow us to propagate low-level acceptability judgments towards the root, while the connections between factored interpretations and choice subsets replicate the logic of selecting a consistent interpretation. Thus, we have satisfied our goal of endowing a single root element with the logic of the entire parse.

Putting it all together, we get a model like the one in Figure 5. User acceptability judgments propagate through the model, determining what choices, choice subsets, and interpretations are compatible with the user's understanding of the sentence. Factored interpretations and the use of a CATMS prevent combinatorial blow-up, while the logical structure of the model ensures that the completeness assumptions (not shown) are faulted when all of their choices are unacceptable.

### 4.3 Problem Decomposition

As discussed above, decomposition allows our algorithm to reformulate the model in response to an initial diagnosis. Decomposition triggers when the inner loop finds a diagnosis that contains a *decomposable fault*, namely one that has an associated decomposition strategy. The outer loop continues until all remaining faults are primitive, meaning there are no strategies to decompose them further. Primitive faults may represent a gap or error in CNLU's knowledge (e.g., "There is a missing semtrans for 'wedge'."), or the closest approximation the algorithm can find (e.g., "There is a missing grammar rule, but it cannot be triangulated further.").

The algorithm is currently greedy. When there are multiple possible decompositions, it picks one arbitrarily. As the set of decomposition strategies grows, we will add support for choosing decompositions that are more likely to succeed and backtracking out of unsuccessful ones.

In the rest of this section, we discuss the three decomposition strategies our system currently supports. So far, we have concentrated our efforts on restrictive semantic errors (C2 and C3 in Table 1). Other error types, in particular grammar errors, will require more complex strategies.

*4.3.1 No Known Semtrans for Word*

This is the simplest decomposition strategy. It triggers automatically when one of the words in the sentence has no known semtranses, as when the system encounters a new word. The only decomposition necessary is to install a completeness assumption for the word's semtranses so it can be faulted. Because all the information the system needs to localize this error is captured in the parse trace, the error can be diagnosed with no user input.

*4.3.2 No Acceptable Semtrans for Word*

Another decomposition strategy handles the case where a word has semtranses, but none of them are applicable to the current sentence. When the completeness assumption for a semantic choice set is faulted, the system walks back over the parse trace, following the choices for the word back to their origins to ensure that no other word senses were ruled out along the way. If this is the case, the system can conclude that a semtrans for the word is missing.

*4.3.3 Missing Valence Pattern for Acceptable Semtrans*

In some cases, CNLU rules out the right semtrans because it is missing a valence pattern. When a grammar rule binds a grammatical role such as the subject or object of a clause, any semtrans that has no way to handle the role (i.e., a valence pattern with an open role of the appropriate type) will be ruled out. This allows us to make a more specific diagnosis than just saying a semtrans is missing. If one of the dropped semtranses would have otherwise applied to the sentence, we can pinpoint the error to a missing valence pattern with the observed grammatical roles.

Because these semtranses were dropped, they do not appear in the model by default. Only when the system has reason to believe a semtrans might have been dropped does it go looking for one, then it installs the necessary structure in the ATMS. The model only contains what it needs.

## 5. Examples

To demonstrate our system's ability to diagnose semantic errors, we applied it to a set of three synthetic errors based on the sentence "Joe ate the apple," which CNLU interprets correctly. For each of the error types our system supports, we ablated a piece of knowledge to induce that error in the sentence. Table 2 shows the examples and the decomposition strategies used for diagnosis.

In all cases, our system is capable of diagnosing the correct error after the user answers only a few natural language questions (or none, when there is no known semtrans).

"apple" only has one semtrans, so removing it means there is no known semtrans for the word. "ate" has two semtranses, `EatingEvent` and `HavingAMeal`, but we only consider the former acceptable. Thus, if we remove the `EatingEvent` semtrans, the system must determine that `HavingAMeal` is unacceptable and diagnose the issue as a missing semtrans. Similarly, removing `EatingEvent`'s valence patterns forces the system to determine whether a semtrans is missing for "ate" or whether the ruled-out `EatingEvent` semtrans is missing a valence pattern.

*Table 2*. Synthetic error examples for "Joe at the apple." The first column shows the knowledge that was removed from CNLU to cause the error. The second shows the error the system diagnosed.

| Ablated Knowledge | Identified Error | ID | Strategy |
|---|---|---|---|
| Semtrans for "apple" | Missing semtrans for "apple" | C3 | No Known Semtrans for Word |
| `EatingEvent` semtrans | Missing semtrans for "eat" | C3 | No Acceptable Semtrans for Word |
| Valence patterns for the `EatingEvent` semtrans | Missing pattern for the `EatingEvent` semtrans | C2 | Missing Valence Pattern for Acceptable Semtrans |

## 6. Related Work

While we are unaware of any existing work that attempts to interactively debug a semantic parser, we draw inspiration from several clusters of related work. The first is past work on model-based diagnosis. de Kleer & Williams (1987) present GDE and demonstrate its use for diagnosing faults in digital circuits. de Koning et al. (2000) diagnose student errors by comparing their responses to a model reasoning trace, using hierarchical decomposition to compactly represent a large search space. Collins (1994) extends traditional model-based diagnosis with fault models generated by a process-centered domain theory, allowing the system to handle novel faults not enumerated by the designer.

More germane to language, *error mining* (de Kok & van Noord, 2017) locates errors in a syntactic parser by running the parser over a corpus and tracking the *n*-grams associated with broken parses. Goodman & Bond (2009) take a similar approach, tracking the combinations of rules that are associated with round-trip parsing and generation failures. These techniques complement INLD nicely, focusing on breadth rather than depth, and future work should explore this synergy.

*Interactive Task Learning* (ITL; Laird et al., 2017) deals with systems that learn tasks through explicit instruction. INLD shares the goal of learning via user interaction, but it attempts to repair or supplement existing linguistic knowledge, rather than teach new tasks. However, ITL systems such as Rosie (Kirk & Laird, 2014) and PUMICE (Li et al., 2019) can learn new vocabulary terms to support their task learning. One can imagine a future system that unifies INLD, ITL, and other forms of interactive learning under the same framework.

Finally, two knowledge-based systems have some intriguing overlap with INLD. KRAKEN (Matthews et al., 2004) and its interactive dialogue component (Witbrock et al., 2003) provide an interface for subject-matter experts to browse, expand, and add entities to the Cyc ontology. The system presented by Kalyanpur (2006) helps users locate and repair errors in OWL ontologies. Both of these systems share INLD's goal of making complex knowledge accessible to inexperienced users, but they handle ontological rather than linguistic knowledge.

## 7. Conclusion

INLD shows promise for making hand-curated language systems more maintainable. In this paper, we have presented a partial implementation of the INLD pipeline for the CNLU semantic parser: an interactive, model-based diagnosis system that locates semantic errors in CNLU's knowledge with the help of the user. We have explained the design of the algorithm, shown how a CNLU parse trace can be converted into a model for diagnosis, and demonstrated our system's ability to diagnose a sample set of synthetic errors.

Notably, our diagnosis system does not require the user to have any expertise aside from a basic knowledge of English. This paves the way for non-experts to help maintain and extend CNLU, which is vital for both longevity of the parser and its use in lifelong learning scenarios (Chen & Liu, 2018).

While the model we have presented here is specific to CNLU, the algorithm can apply to any language system that relies on hand-curated linguistic resources, such as the NLU component of OntoAgent (McShane & Nirenburg, 2021) or the English Slot Grammar parser in Watson (McCord, Murdock, & Boguraev, 2012). Formulating a model for a new system is non-trivial, but the basic pattern of completeness assumptions, acceptability judgments, and decomposition strategies should hold for a wide range of systems.

## 8. Future Work

While the system described in this paper constitutes a baseline interactive diagnosis module for CNLU, much work remains to be done, both for NL diagnosis and for INLD as a whole. For the purposes of this paper, we have circumscribed the task of diagnosis, focusing on the classes of errors that are most amenable to INLD. Future work will explore other categories of errors, more sophisticated debugging strategies, and the remaining two stages of the pipeline.

We observe that the current limiting factor on our system is its ability to generate English paraphrases of NextKB concepts. While the canned strings and generation templates stored in NextKB have broad coverage, they are incomplete. Furthermore, they have never been tested in a scenario as strenuous as this. Paraphrases that are perfectly fine in isolation may be confusing when used to disambiguate related concepts. Further work is needed to both expand coverage for generation and explore strategies for clarifying paraphrases. This will enable us to test how well our diagnosis strategies work on wild text.

One major diagnosis strategy discussed in Nakos et al. (2022) but omitted here is differential diagnosis, the strategy of rephrasing a sentence to localize the error. By rephrasing a sentence, parsing it, and comparing the user's acceptability judgments for the two sentences, INLD can localize the error to either the part of the sentence that changed or the part that stayed the same. In this way, it can drill down on errors that might not be revealed by other localization strategies and break up complex sentences into manageable parts. Implementing this strategy will require

INTERACTIVELY DIAGNOSING ERRORS IN A SEMANTIC PARSERreliable support for simplification, substitution, and comparison between parses, all of which will contribute to a robust INLD system.

## Acknowledgements

Thanks to the anonymous reviewers for their helpful feedback. This research was sponsored by the US Office of Naval Research under grant #N00014-20-1-2447.## References

Allen, J. F. (1994). *Natural language understanding*. (2nd ed). Redwood City, CA: Benjamin/Cummings.

Chen, Z., & Liu, B. (2018). *Lifelong machine learning*. (2nd ed). Morgan & Claypool Publishers.

Collins, J. W. (1994). *Process-based diagnosis: An approach to understanding novel failures*. Doctoral dissertation, Department of Computer Science, University of Illinois at Urbana-Champaign, Urbana, IL.

Crouse, M. (2021). *Question-answering with structural analogy*. Doctoral dissertation, Department of Computer Science, Northwestern University, Evanston, IL.

de Kleer, J. (1986). An assumption-based TMS. *Artificial Intelligence*, 28(2), 127-162.

de Kleer, J., & Williams, B. C. (1987). Diagnosing multiple faults. *Artificial Intelligence*, *32*(1), 97-130.

de Kok, D., & van Noord, G. (2017). Mining for parsing failures. In M. Wieling, M. Kroon, G. van Noord, & G. Bouma (Eds.), *From Semantics to Dialectometry: Festschrift in honor of John Nerbonne.* College Publications.

de Koning, K., Bredeweg, B., Breuker, J., & Wielinga, B. (2000). Model-based reasoning about learner behaviour. *Artificial Intelligence*, *117*(2), 173-229.

DeCoste, D. and Collins, J. W. (1991). *CATMS: An ATMS which avoids label explosions* (Technical report). Northwestern University, Evanston, IL.

Fillmore, C. J., Wooters, C., and Baker, C. F. (2001). Building a large lexical databank which provides deep semantics. *Proceedings of the 15th Pacific Asia Conference on Language, Information and Computation* (pp. 3-26). Hong Kong, China.

Forbus, K. D., Hinrichs, T. (2017). Analogy and qualitative representations in the Companion cognitive architecture. *AI Magazine*, *38*: 34-42.

Goodman, M. W., & Bond, F. (2009). Using generation for grammar analysis and error detection. *Proceedings of the ACL-IJCNLP 2009 Conference Short Papers* (pp. 109-112).

Hinrichs, T., & Forbus, K. (2014). X goes first: Teaching a simple game through multimodal interaction. *Advances in Cognitive Systems*, *3*, 31-46.

Kalyanpur, A. (2006). *Debugging and repair of OWL ontologies*. Doctoral dissertation, Department of Computer Science, University of Maryland, College Park, MD.

Kirk, J. R., & Laird, J. E. (2014). Interactive task learning for simple games. *Advances in Cognitive Systems*, 3, 13-30.15